\documentclass[conference]{IEEEtran}
\usepackage{graphicx}
\usepackage{hyperref}
\usepackage{aux/macros}

\def\rhomax{\rho_{\sf max}}
\renewcommand{\x}{x}
\providecommand{\thetavec}{\bm{\theta}}
\providecommand{\loss}{\mathsf{L}}
\providecommand{\Prob}{\mathbb{P}}
\providecommand{\Xt}{\widetilde{X}}
\renewcommand{\X}{X}

\providecommand{\xt}{\widetilde{x}}
\providecommand{\pt}{\widetilde{p}}

\def\citet{\cite}
\usepackage{apptools}
\crefname{subappendix}{\IfAppendix{section}{appendix}}{\IfAppendix{sections}{appendices}s}

\begin{document}

\makeatletter
\newcommand{\linebreakand}{%
  \end{@IEEEauthorhalign}
  \hfill\mbox{}\par
  \mbox{}\hfill\begin{@IEEEauthorhalign}
}
\makeatother

\title{Local Convergence of Gradient Descent-Ascent for Training Generative Adversarial Networks}

\author{\IEEEauthorblockN{Evan Becker}
\IEEEauthorblockA{CS, UCLA\\
\texttt{evbecker@ucla.edu}}
\and
\IEEEauthorblockN{Parthe Pandit}
\IEEEauthorblockA{HDSI, UCSD \\
\texttt{parthepandit@ucsd.edu}}
\and
\IEEEauthorblockN{Sundeep Rangan}
\IEEEauthorblockA{ECE, NYU\\
\texttt{srangan@nyu.edu}}
\and
\IEEEauthorblockN{Alyson K. Fletcher}
\IEEEauthorblockA{Statistics, UCLA \\
\texttt{akfletcher@ucla.edu}}
}

\maketitle

\begin{abstract}
Generative Adversarial Networks (GANs) are a popular formulation to train generative models for complex high dimensional data. The standard method for training GANs involves a gradient descent-ascent (GDA) procedure on a minimax optimization problem.  This procedure is hard to analyze in general due to the nonlinear nature of the dynamics.
We study the local dynamics of GDA for training a GAN with a kernel-based discriminator. This convergence analysis is based on a linearization of a non-linear dynamical system that describes the GDA iterations, under an \textit{isolated points model} assumption from \cite{becker2022instability}. Our analysis brings out the effect of the learning rates, regularization, and the bandwidth of the kernel discriminator, on the local convergence rate of GDA. Importantly, we show phase transitions that indicate when the system converges, oscillates, or diverges. We also provide numerical simulations that verify our claims. 
\end{abstract}

\thispagestyle{plain}
\pagestyle{plain}
\section{Introduction}
Modelling complex signals such as images, speech, text is of broad interest in machine learning and signal processing. Generative models for such data can enable many engineering and scientific applications such as sampling, inference, and understanding structural properties of complex data. With the increasing access to computational resources, the recent focus of generative modelling has been using data-driven techniques.

Generative Adversarial Networks (GANs) are a class of probabilistic generative models that avoid expensive likelihood computations while still providing good sample quality \citep{goodfellow2014generative}. In order to fit complex data distributions, two models (typically deep neural networks) are trained in an alternating manner: a \emph{generator} $\mathcal{G}$ learns a deterministic map from a latent space $\mc Z$ to the data space $\mc X$, while a \emph{discriminator} or \emph{critic} model $\mathcal{D}$ attempts to discern whether a sample belongs to the training dataset or the generated dataset. 

The discriminator plays an important, yet poorly understood, role in the training of a GAN. It is well known from \citep{goodfellow2014generative} that if the discriminator is trained to minimize the cross-entropy between true and generated samples, the generator would minimize the Jensen-Shannon divergence between the true distribution and the distribution of the generated samples. Similarly different choices for discriminator loss functions lead to a variety of $f$-divergences \cite{nowozin2016f} and probability metrics \cite{gretton2012kernel} between the generated and true distributions.

In practice, however, we apply GDA for training GANs, whereby the discriminator is not allowed to converge, making analysis of the iterative training extremely difficult. Furthermore, training commonly suffers from empirical breakdowns such as mode collapse, in which the entire generated distribution converges to a small portion of the target distribution \cite{salimans2016improved}. The generator may even fail to converge entirely when gradients from the discriminator are too small for the generator to proceed during training. Without an understanding of when and how these phenomena occur, practitioners have to rely on heuristics and extensive hyperparameter tuning based on trial and error procedures \cite{salimans2016improved,karras2017progressive,miyato2018spectral}. 

In this work, we characterize the local convergence rates of GAN training when the discriminator is kernel-based. This choice of the discriminator model is motivated by the recently discovered equivalence between wide neural networks and kernel methods via the Neural Tangent Kernel framework \cite{jacot2018neural}. While the discriminators problem is simplified due to the kernel-based discriminator, the overall dynamics of the generated samples remain non-linear and complex, and hence retain many of the properties exhibited by GANs in practice such as mode collapse and divergence \cite{becker2022instability}.

\subsection{Prior Work on Linear GANs and Main Contributions}
Stability analysis for GANs under stylized settings goes back to the Dirac-GAN framework from \cite{mescheder2018training}, which looked at the local stability of a two-point system using a linear discriminator to demonstrate examples of catastrophic forgetting. Other GAN works use a similar linearization analysis, such as \citet{nagarajan2018gradient, mroueh2021convergence}. The isolated points model proposed by \cite{becker2022instability}
allowed for a more complex model while remaining analytically tractable, by letting the generated probability mass differ from the true mass in various isolated regions. We provide new insight into the framework proposed by \cite{becker2022instability} by going beyond stability analysis and characterizing rates of convergence.

We analyze the local convergence of the non-linear dynamical system that describes the GDA iterates, in settings when the equilibrium is stable. Our analysis is based on a linearization of these non-linear dynamics.
We show how changing the kernel width can improve the rate of convergence, while also highlighting a phase transition under which the convergence remains unaffected by changes in the kernel width.

\section{Model}\label{sec:model}
We investigate the training dynamics of a GAN where the target distribution and the generated distribution are discrete point masses, following the framework of  \citep{becker2022instability}. We highlight key elements of our model below.
\subsection{Target and Generated Distributions} We assume that the target and generated distributions consist of point masses with density functions over $x\in\Real^d$ given by
\begin{equation} \label{eq:probrg}
    \Prob_r(\x) = \sum_{i=1}^{N_r} p_i \delta(\x-\x_i), 
    \quad
    \Prob_g(\x) = \sum_{j=1}^{N_g} \pt_j \delta(\x-\xt_j), 
\end{equation}
where $\delta$ is the Dirac delta function, $\X=\set{\x_i}_{i=1}^{N_r}$ and $\wt\X=\set{\xt_j}_{j=1}^{N_g}$ are the true and generated points, and 
$\set{p_i}_{i=1}^{N_r}$ and $\set{\pt_j}_{j=1}^{N_r}$ are their (fixed) probability masses.
The problem we consider is learning the locations $\Xt$
so that the generated and true distributions match. Thus the decision variable of the generator model is $\wt \X.$ This simplification is justified since we wish to study the role of the discriminator in this work.


\subsection{Kernel Discriminator} The GAN discriminator
is a function $f:\mc X\to\Real$ which predicts whether a sample $\x$ is real or fake based on $\sign(f(\x))$. In this paper we assume that the discriminator belongs to a Reproducing Kernel Hilbert Space $\Hilbert$ corresponding to a bivariate positive definite kernel function $K:\Real^d\times\Real^d\to\Real.$
The discrimininator defines a maximum mean discrepancy (MMD) metric
\begin{align}\label{eq:mmd}   \textsf{MMD}(\Prob_r,\Prob_g):=\max{\substack{f\in\Hilbert\\\norm{f}\leq 1}} \sum_{i=1}^{N_r} p_i f(x_i) - 
    \sum_{i=1}^{N_g} \wt p_i f(\wt x_j)
\end{align}
between $\Prob_r$ and $\Prob_g,$ which the generator tries to minimize.

\subsection{Minimax Optimization Formulation for Training a GAN}  
We assume a mini-max loss function similar to \citep{arjovsky2017principled, becker2022instability, li2017mmd, unterthiner2017coulomb}
of the form:
\begin{subequations}\label{eq:minmax}
\begin{align} 
&\min{\wt\X}\max{f\in\Hilbert}\loss(f,\wt \X)\\
&\loss(f,\wt\X) := \sum_{i=1}^{N_r} p_i f(x_i) - 
    \sum_{i=1}^{N_g} \wt p_i f(\wt x_j) - \tfrac{\lambda}{2}\norm{f}_\Hilbert^2.\label{eq:loss}
\end{align}
\end{subequations}
The regularization parameter $\lambda>0$ is some constant, that acts as a Lagrange multiplier for the optimization problem in \cref{eq:mmd}. 
The loss is a function of the discriminator $f$
and generated samples $\wt \X$.

\textbf{Notation:} For matrices $X\in\Real^{n\times d}$, $Z\in\Real^{p\times d}$ with rows $x_i,z_j\in\Real^d$, and for vectors $u,v,x,z\in\Real^d$, and a kernel function $K:\Real^{d}\times\Real^{d}\to\Real$,
by $K(X,Z)$ we denote the $n\times p$ matrix with $ij^{\rm th}$ entry $K(x_i,z_j)$. By $\grad_1 K(x,z)$ we denote the map $\Real^{d}\times\Real^d\to\Real^d$ given by $(u,v)\mapsto\frac{\partial}{\partial x}
K(x,z)\big|_{x=u,z=v}.$  Similarly, by $\grad_1 K(X,Z)\p$ we denote the $n\times d$ matrix with $i^{\rm th}$ row $ \sum_{i=1}^p \grad_1 K(x_i,z_j)p_j$. Furthermore for a vector $\v\in\Real^n$ and matrix $\M\in\Real^{n\times d}$, by $\v\odot \M$ we denote the Hadamard product, which yields a $n\times d$ matrix with $ij^{\rm th}$ entry $v_i\M_{ij}$. For example, $\v\odot\grad_1K(X,Z)\p$ is a $n\times d$ matrix with $i^{\rm th}$ row $v_i\frac{\partial}{\partial x}\round{ \sum_{j=1}^p K(x,z_j)p_j}\big|_{x=x_i}$

\subsection{Training Dynamics of Gradient Descent Ascent}

We assume the generator performs gradient descent on the above minimax optimization problem with a step size $\eta_g$ 
and the discriminator performs gradient ascent with step size $\eta_d$. We let $(f^t,\Xt^t)$ denote the discriminator and generated samples
in step $t$.
\begin{subequations} \label{eq:gda_def}
\begin{align}
    f^{t+1} &= f^{t} + \eta_d\frac{\partial}{\partial f}\loss(f^t, \Xt^t)\\
    \Xt^{t+1} &= \Xt^{t} - \eta_g \frac{\partial}{\partial \Xt}\loss(f^t,\Xt^t) 
\end{align}
\end{subequations}
where the first equation uses the Fr\'echet derivative with respect to $f$. This can be simplified since the loss function $\loss(f,\wt X)$ only consists of linear and quadratic terms of $f.$ Observe that for any $u$ in $\Real^d$, the linear term $f(u)=\inner{f,K(u,\cdot)}_\Hilbert$ due to the reproducing property of the kernel whereby, $\frac{\partial}{\partial f}f(u)$ is the function $v\mapsto K(u,v),$ denoted $K(u,\cdot).$

Using the loss function in \cref{eq:minmax}, we get the updates
\begin{subequations}\label{eq:gda}
\begin{align}\label{eq:gda_f}
    f^{t+1} &=  (1-\lambda\eta_d)f^t  +  \eta_d\round{ K(\cdot,\X)\p  - K(\cdot,\wt \X_i)\wt \p}\\
    \xt_j^{t+1} &= \xt_j^{t} + \eta_g\wt p_j \grad f^t(\wt x_j^t),\qquad \forall\ j=1,2,\ldots,N_g
\end{align}
\end{subequations}

Notice that \eqref{eq:gda_f} is linear in $f$, whereby we can simplify these equations further. The following lemma simplifies \cref{eq:gda} by eliminating the discriminator $f.$
\begin{lemma}[Training Dynamics] Assume $f_0 = 0,$ the zero function in the RKHS $\Hilbert.$ Then the following deterministic dynamical system describes the evolution of the samples generated using \cref{eq:gda}.
\begin{align}\label{eq:gda_X}
    \wt\X^{t+1}=&\wt\X^t + \eta_d\eta_g\sum_{s=0}^t (1-\lambda\eta_d)^{t-s}\times\nonumber\\
    &\wt\p\odot\round{\grad_1 K(\wt\X^t,\X)\p-\grad_1 K(\wt\X^t,\wt\X^s)\wt\p}
\end{align}
\end{lemma}
Note that the above dynamical system is nonlinear in $\Xt$, and is non-Markovian due to dependence of $\Xt^{t+1}$ on $\set{\Xt^{s}}_{s\leq t}$. The term $\wt\p\odot\grad_1 K(\wt \X,\X)\p$ can be thought of as a \textit{drift}, whereas $\wt\p\odot\grad_1 K(\wt \X,\wt \X)\wt\p$ can be thought of as a \textit{diffusion}.

From \cref{eq:gda_X} we can immediately infer a condition for a set of generated points $\wt X^*$ to be in equilibrium
\begin{lemma}
    A set of points $\wt X^*$ is in equilibrium for the dynamics \cref{eq:gda_X} if and only if
\begin{align}\label{eq:sufficient_eq}
    \grad_1 K(\wt\X^*,\X)\p=\grad_1 K(\wt\X^*,\wt\X^*)\wt\p.
\end{align}
\end{lemma}
\begin{remark}
    The set of equilibrium points depend only on the kernel $K$ and are invariant to the hyperparameters $\eta_d,\eta_g,\lambda.$ However the dynamics and convergence properties of these equilibria depend on $\eta_d,\eta_g,\lambda$ as well as $\sigma$.
\end{remark}

\subsection{Model and Optimization Hyperparameters}
Our analysis characterizes the effect of four hyperparameters in total which can be categorized as modelling and optimization hyperparameters. 

This setting has two model hyperparameters that control the smoothness of the discriminator. The regularization $\lambda$ controls the $\Hilbert$-norm of the discriminator, which is a measure of global smoothness. In contrast, the kernel bandwidth $\sigma$ is a measure of the local smoothness. We also have two optimization hyperparameters, which are the learning rate of the generator $\eta_g$ and the learning rate of the discriminator $\eta_d$. In practice often $\eta_g\ll\eta_d.$

\section{Local Convergence around True Samples}

\subsection{Assumptions on the kernel function}
We assume the the kernel $K(\x,\x')$ is smooth and,
at each true point $\x_i$:
\begin{subequations}\label{eq:k_as}
\begin{align}\label{eq:kgrad_as}
    &\grad_1 K(x_i,x_i)=\left. \frac{\partial K(\x,\x_i)}{\partial \x}
    \right|_{\x=\x_i} = \zero,\\
 \label{eq:khess_as}
    &\left. -\frac{\partial^2 K(\x,\x_i)}{\partial \x^2}
    \right|_{\x=\x_i} =
    \left. \frac{\partial^2 K(\x,\x')}{\partial \x \partial \x'}
    \right|_{\x=\x'=\x_i} = \frac{1}{\sigma^2}\I_d
\end{align}
\end{subequations}
for some $\sigma > 0$ that we call the \emph{kernel width} and represents
the curvature of the kernel around $\x = \x_i$. Note that 
\eqref{eq:kgrad_as} and \eqref{eq:khess_as} are satisfied for the standard
RBF kernel:  
\begin{equation} \label{eq:krbf}
    K(\x,\x') = \exp\left( - \tfrac{1}{2\sigma^2}\|\x-\x'\|^2
    \right).
\end{equation}
\begin{proposition}\label{prop:sufficient_eq}
    Under the above assumption, $\Xt^*$ such that $\wt x_j^*=x_i$ for some $i$, is an equilibrium.
\end{proposition}
This follows immediately from the observation in \cref{eq:sufficient_eq}. When the assumptions on the kernel \cref{eq:k_as} are satisfied, both sides of \cref{eq:sufficient_eq} vanish.

The results in \cite{becker2022instability} analyzed the stability of this equilibrium under an isolated points model described below, which localizes the analysis around each true point.

\subsection{Isolated Points Model}
We assume the true samples are separated 
far enough so that there exists a non-empty \textit{isolated neighborhood} 
$V_i$ around each sample $\x_i$ such that,
\begin{equation}  \label{eq:Kdist}
    K(\x,\x') = 0 \mbox{ for all } \x \in V_i 
    \mbox{ and } \x' \in V_j \text{ for all $i\neq j$}.
\end{equation}
In other words, the generated points are separated sufficiently far apart such that they are outside the width of the kernel evaluated at another sample.
We let $\mc N_i$ be the set of indices $j$ such that the generated  
points $\xt_j^t \in V_i$, for all $t$. 

Thus the dynamics we study can be written as
\begin{align}\label{eq:gda_X_local}
    \wt\X^{t+1}_i=&\wt\X^t_i + \eta_d\eta_g\sum_{s=0}^t (1-\lambda\eta_d)^{t-s}\times \nonumber\\
    &\wt\p_i\odot\round{\grad_1 K(\wt\X^t_i,x_i)p_i-\grad_1 K(\wt\X^t_i,\wt\X^s_i)\wt\p_i}
\end{align}
where $\wt\X_i^t$ are points generated inside the region $V_i,$ and $\wt\p_i$ is the length $\abs{\mc N_i}$ subvector of $\wt p$ corresponding to these points.

Under this assumption, if $j\in \mc N_i$ and $k\notin \mc N_i$, then \cref{eq:gda_X_local} ignores interaction terms between $(\wt x_k,x_i)$, and $(\wt x_j,\wt x_k)$, compared to \cref{eq:gda_X}. Note \cref{eq:gda_X} tracks $N_r N_g + N_g^2$ interaction terms whereas \cref{eq:gda_X_local} only tracks $|\mc N_i|+|\mc N_i|^2$ terms where $|\mc N_i|$ is the number of generated points inside $V_i.$

We will call the updates \cref{eq:gda_X_local}
the \emph{dynamical system in the region $V_i$}. For the purpose of analysis it is beneficial to write the dynamics involving both the discriminator and the generator as below:
\begin{subequations}\label{eq:gda_local}
\begin{align}
    f^{t+1}_i &= (1-\lambda\eta_d)f^t +  K(\cdot,x_i)p_i - \sum_{i\in \mc N_i} K(\cdot,\wt x_i)\wt p_i \\
    \xt_j^{t+1} &= \xt_j^{t} + \eta_g\wt p_j \grad f^t(\wt x_j)\qquad x_j\in V_i
\end{align}
\end{subequations}
Under the isolated points model, the discriminator satisfies 
\begin{subequations}
\begin{align}
    f^t(u) &= \sum_{i=1}^{N_r}f_i^t(u),\qquad \forall\,u\\
    f^t(x)&=f^t_i(x)\qquad\forall\, x\in V_i.
\end{align}
\end{subequations}

\subsection{Main result}
Given a local region $V_i$,
we wish to study the dynamics of the local system given by \cref{eq:gda_local}
where $\xt_j$ are close to $\x_i$
for all $j \in \mc N_i$.  That is, all the generated points are close to the true
point in that region.
To this end, we write the local updates \eqref{eq:gda_local} as a mapping
\begin{equation} \label{eq:Phidef}
    (f^{t+1}_i,\Xt_i^{t+1}) = \Phi_i(f^{t}_i,\Xt_i^{t}),
\end{equation}
where $\Phi_i(\cdot)$ represents the update function in \eqref{eq:gda_local}.
Also, let
\begin{equation} \label{eq:xeq}
    \Xt^*_i = \{ \xt^*_j,~ j \in\mc N_i \}, \quad \xt^*_j = \x_i.
\end{equation} 
It is shown in \citep{becker2022instability} that there exists a parameter vector
$f_i^*$ such that $(f_i^{*},\Xt_i^{*})$ is an \emph{equilibrium point} of
$\Phi_i(\cdot)$ in that
\begin{equation} \label{eq:xteq}
    (f_i^{*},\Xt_i^{*}) = \Phi_i(f_i^{*},\Xt_i^{*}).
\end{equation}
The condition \eqref{eq:xteq} implies that if $(f_i^{t},\Xt_i^{t})
= (f_i^{*},\Xt_i^*)$ for some $t$, then $(f_i^{s},\Xt_i^{s})$
will remain at $(f_i^{*},\Xt_i^*)$ for all subsequent times $s \geq t$.
Let $\J^*_i$ denote the Jacobian of the update mapping $\Phi_i(\cdot)$
at the equilibrium point $(f_i^{*},\Xt_i^{*})$ and define the spectral radius of the Jacobian
\begin{equation} \label{eq:rhomax}
    \rhomax:=\rhomax(\J^*_i) = \textsf{max}\set{ |\rho| \mid \rho \in \mathrm{spec}(\J^*_i)},
\end{equation}
where $\mathrm{spec}(\J^*_i)$ is the spectrum of $\J^*_i$, i.e., its eigenvalues.

A well-known result of non-linear systems theory  \citep{vidyasagar2002nonlinear} is that
the equilibrium point $(f_i^{*},\Xt_i^*)$ is \emph{locally asymptotically stable}
if $\rhomax(\J^*_i)< 1$.  
Conversely, if $\rhomax(\J^*_i) > 1$, the system can be shown to 
be \emph{locally unstable} -- see \citep{vidyasagar2002nonlinear} for precise definitions.
Hence, $\rhomax(\J^*_i)$ can provide necessary and sufficient conditions for local stability.
Also, if $\rhomax< 1$ and the system is initialized 
at $(f_i^{0},\Xt_i^0)$ sufficiently close to $(f_i^{*},\Xt_i^*)$ then, 
the components will converge geometrically as
\begin{equation} \label{eq:convratelmax}
    \| \xt_j^t - \xt_i \| \leq C \rhomax^t\| \xt_j^0 - \xt_i \|,
\end{equation}
for some constant $C$.  
Hence, the spectral radius $\rhomax$ also provides a measure of the convergence
rate of the system. Our main theorem below applies this result to obtain an exact characterization of the convergence rate of the local dynamics by studying the spectrum of $\J^*$ in terms of the model and optimization hyperparameters. 

Recall the model hyperparameters: $\sigma$ -- discriminator kernel width, $\lambda$ -- IPM regularization, $\eta_d$ -- learning rate of discriminator and $\eta_g$ -- learning rate of the generator.



\begin{figure*}
    \centering
\includegraphics[width=\textwidth]{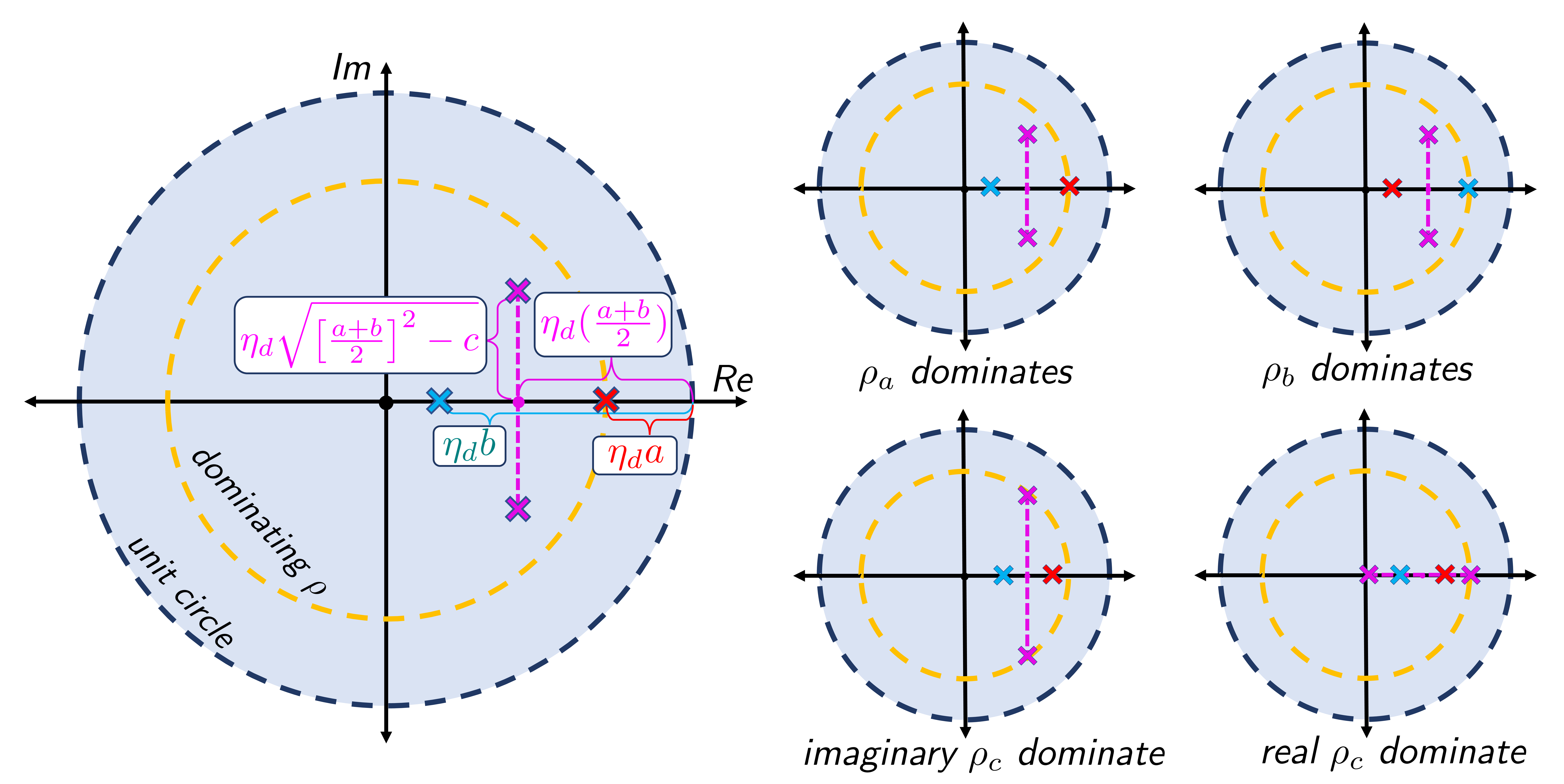} \caption{Eigenvalues of the linearized system in \Cref{thm:discrete}. Here $a:=\lambda$ the regularization, $b:=\frac{\mu\pt\Delta_i}{\lambda\sigma^2}, 
    $ and $
    c:=\frac{\mu\pt p_i}{\sigma^2},$ where $ \mu := \frac{\eta_g}{\eta_d}$,  where $\sigma$ is the kernel width, and $\eta_d$ and $\eta_g$ are learning rates of the discriminator and generator respectively.}
    \label{fig:linear-eigenvalues}
\end{figure*}

\begin{theorem} \label{thm:discrete}
Consider the isolated neighborhood training dynamics in \eqref{eq:gda_local} under the assumptions in Section~\ref{sec:model} in some region $V_i$. 
Additionally, assume that the weights of the generated points are equal so that $\pt_j = \pt$ for some $\pt > 0$ and all $j \in N_i$.  Define
\begin{equation} \label{eq:abcdef}
    a:=\lambda, \quad
    b:=\frac{\mu\pt\Delta_i}{\lambda\sigma^2}, 
    \quad
    c:=\frac{\mu\pt p_i}{\sigma^2}, \quad \mu := \frac{\eta_g}{\eta_d}, 
\end{equation}
and 
\begin{equation} \label{eq:deldef}
     \Delta_i := p_i - \sum_{j \in\mc N_i} \pt_j = p_i - |\mc N_i|\pt.
\end{equation}
Then, 
the eigenvalues of the $\J^*$ are
of the form
\begin{equation} \label{eq:rhonu}
    \rho = 1-\eta_d \nu,
\end{equation}
where $\nu$ is from the set:
\begin{align} \label{eq:nuset}
    \nu \in \begin{cases}
        \set{a, b, m \pm \sqrt{m^2-c}}
        & \text{if } |\mc N_i| > 1 \\
        \set{a, m \pm \sqrt{m^2-c}}
        & \text{if } |\mc N_i| = 1.
        \end{cases}
\end{align}
where $m = (a+b)/2$.
\end{theorem}

The proof of the result is given in Appendix~\ref{sec:linearized-rate-proof} and
builds on the linear analysis in \citep{becker2022instability}.  The theorem above gives an exact characterization of the eigenvalues of the linear system in terms
of the key parameters including the step sizes
and kernel width.

\subsection{Selecting the step size}
An immediate consequence of \Cref{thm:discrete} is that it guides the selection of the step-sizes that ensure local stability.  As described above, for local stability, we wish that $|\rho| < 1$
for all $\rho$ in \eqref{eq:rhonu}.  
The following provides necessary and sufficient conditions 
on $\eta_d$ for this stability condition to occur.

\begin{corollary}\label{cor:etad-constraint}  Under the conditions in \Cref{thm:discrete},
the spectral radius of the Jacobian, $\rhomax(\mathrm{spec}(\J^*_i)) < 1$, if and only if:
\begin{align} \label{eq:etad_ub1}
        0 < \eta_d < \begin{cases} \text{min}\left\{ \frac{2}{a},\: \frac{2}{b},\: \frac{a+b}{c} \right\} & \mbox{if } |\mc N_i| > 1 \\
        \text{min}\left\{ \frac{2}{a},\: \frac{a+b}{c} \right\} & \mbox{if } |\mc N_i| = 1.
        \end{cases}
\end{align}
\end{corollary}
In particular, by choosing $\eta_d$ small enough, we can always guarantee the system is locally stable when $\lambda > 0$ 
and $\Delta_i > 0$, meaning that there is at least some
regularization and the true point mass exceeds the generated
point mass.  We can also derive a simple sufficient condition:

\begin{proposition}[Sufficient condition for stability]\label{prop:linear-sufficient-condition}
The training dynamics \cref{eq:gda_local} are stable around equilibrium $\wt\X_i^*$ from \cref{eq:xeq} for all $\Delta_i \in (0,p_i)$ if,
\begin{align}\label{eq:stable_lr}
    \eta_d < \frac2\lambda,\qquad\text{and}\qquad\eta_g<\lambda\sigma^2.
\end{align}
\end{proposition}

The rest of the paper assumes \cref{eq:stable_lr} holds and derives convergence rates based on the choice of kernel width $\sigma.$

\section{Convergence Rate 
and Kernel Width}
\begin{figure}[t]
\includegraphics[width=\columnwidth]{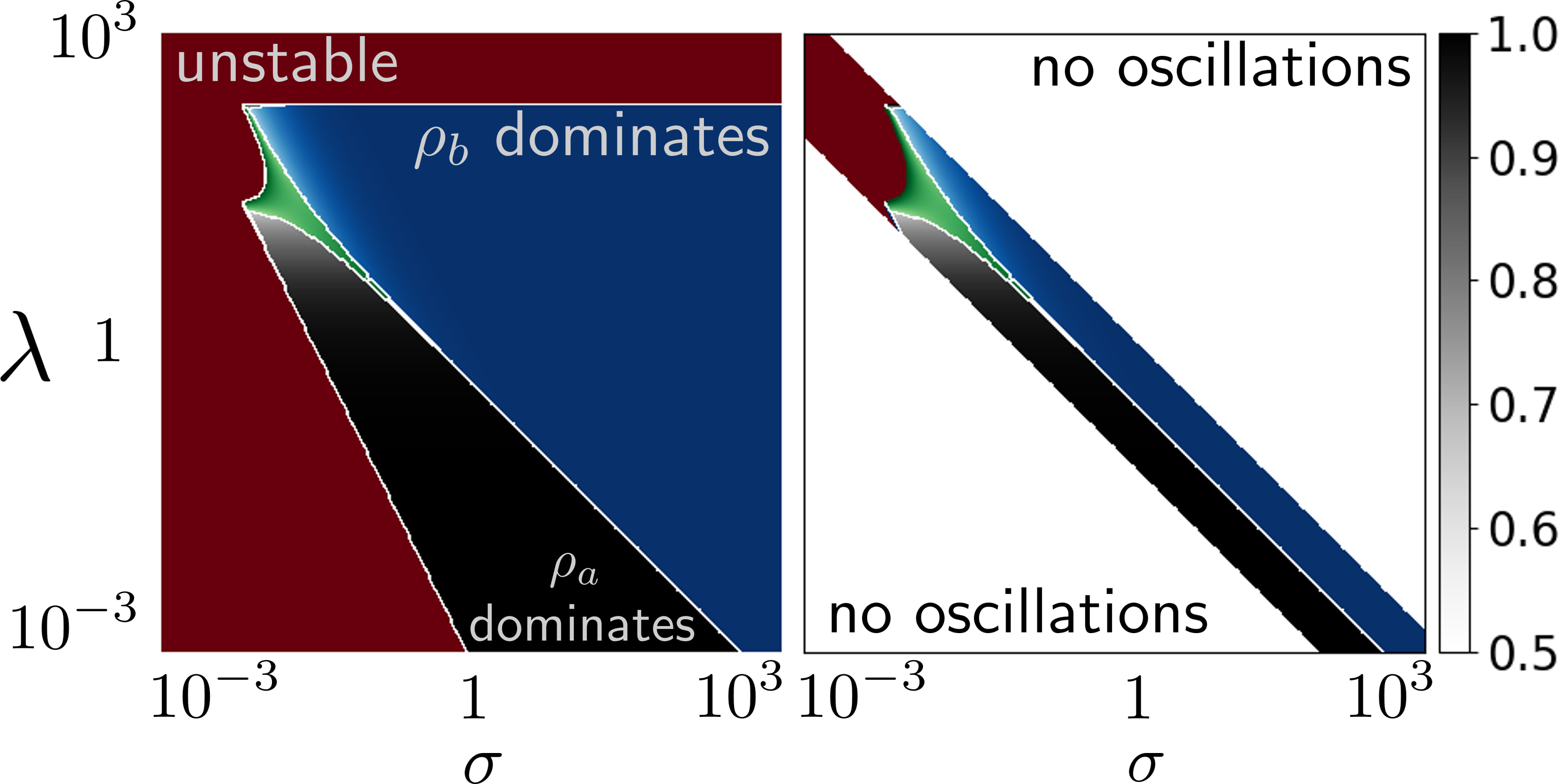}
\caption{(Left) Heat map of $|\rho_{\sf max}|^2$ as a function of $(\sigma,\lambda)$, the kernel width and regularization. The colorbar shows the intensity of $\rhomax.$ Learning rates are $\eta_d=\eta_g=1e^{-2}$, $p_i=1$ and $\pt=0.8$ whereby $\Delta_i=0.2, \mu=1$. The region for which $|\rho_{\sf max}|^2>1$ is highlighted in red, where $\rho_a$ dominates is in black, $\rho_b$ dominates is in blue and where $\rho_c$ dominates is in green. In the black region where $\rho_a$ dominates, $\rhomax$ is insensitive to $\sigma.$ (Right) The region in which $\rho_c$ is complex is selected from the heat map on the left. In this region the system shows oscillatory behavior due to the imaginary part of $\rho_c$.}
    \label{fig:phase_transition}
\end{figure}
\Cref{thm:discrete} can also provide insights into
the relation of the convergence rate to the 
system parameters.
Specifically, recall  from 
\cref{eq:convratelmax} that the spectral
radius, $\rhomax(\J^*_i)$
defined in \cref{eq:rhomax},
determines the convergence rate of the local dynamics, i.e., $\rhomax$ closer to 1 indicating slower convergence and $\rhomax$ closer to 0 indicates faster convergence. 
Now, among the values in \eqref{eq:rhonu}, 
the $\rho$ that maximizes
$|\rho|$  will be one of three values:
\begin{equation}  \label{eq:rhoabc}
    \rho_a = 1-\eta_d a, \ 
    \rho_b = 1-\eta_d b, \ 
    \rho_c = 1-\eta_d ( m   - \sqrt{m^2 - c} 
    ),    
\end{equation}
where $m = (a+b)/2$.  The cases when the different values dominate are shown in \Cref{fig:linear-eigenvalues}.

It is clear from \cref{eq:rhoabc} that controlling the dominant eigenvalue by adjusting the relevant hyperparameters can improve the rate of convergence.

\begin{remark}[Saturation with Kernel Width] We now share a phase of the dynamical system where changing $\sigma$ does not affect the convergence rate.
Consider the dynamics \eqref{eq:gda_local} with fixed $\pt$, and $|\mc N_i|$. Furthermore, assume $\eta_d$ is fixed such that the condition from Corollary \ref{cor:etad-constraint} is satisfied. Then changing the kernel width parameter $\sigma^2$ cannot improve the convergence rate in the following settings:
\begin{itemize}
    \item When all eigenvalues are real and $\rho_a$ dominates. This condition is equivalent to $c<\frac{1}{4}(a+b)^2$ and $a<\min{} \left(b, m-\sqrt{m^2-c}\right)$. 
    \item When two eigenvalues are complex and $\rho_a$ dominates. Equivalently $c>\frac{1}{4}(a+b)^2$ and $a<\text{min}\set{b, 2m-\eta_d c}$.
\end{itemize}\label{rem:saturation}
\end{remark}

\subsection{Diminishing learning rate}
One example regime in which this saturation can clearly be understood is when the learning rate is small, $\Delta_i$ is positive, and $\rho_c$ is complex.
When the learning rate is small, the magnitude for any eigenvalue of the form $\rho_\nu = 1-\eta_d \nu$ can be approximated by $|\rho_\nu|^2 \approx 1-2\eta_d \text{Re}(\nu) +\mathcal{O}(\eta_d^2)$. This means that we have eigenvalues with approximate magnitudes:
\begin{subequations}
\begin{align}
    |\rho_a|^2 &\approx 1-2\eta_d\lambda \\
    |\rho_b|^2 &\approx 1-2\eta_d\frac{\mu\pt \Delta_i}{\lambda \sigma^2} \\
    |\rho_c|^2&\approx 1-\eta_d(\lambda + \frac{\mu\pt \Delta_i}{\lambda \sigma^2})
\end{align}
\end{subequations}
This yields that the largest eigenvalue is
\begin{align}
    1-2\eta_d\cdot\text{min}\set{\lambda,\frac{\mu\pt \Delta_i}{\lambda \sigma^2}}
\end{align}
Thus reducing the kernel width $\sigma$ below $\sqrt{\mu\wt p\Delta_i}/\lambda$, does not lead to changes in the convergence rate $1-2\eta_d\lambda$.

\section{Numerical Results}
In this section, we demonstrate the accuracy of our linearized dynamics by comparing predicted convergence to actual GAN training behavior around local equilibrium. 

\paragraph{Phase transitions} In \Cref{fig:phase_transition}, we plot the heatmap of dominating eigenvalue magnitude for a range of regularization and kernel width settings. Note that in this figure we use small learning rate ($\eta_d = \eta_g = 1e^{-2}$), meaning firstly that the system is stable for almost all choices of hyperparameters (\Cref{fig:phase_transition}a). In the middle plot (\Cref{fig:phase_transition}b), it can be observed that the majority of fast convergence behaviors occur when $\rho_c$ has imaginary components. In order to analytically find this region, the condition $m^2<c$ provides a quadratic inequality in terms of $\gamma=1/\sigma^2$, from which the roots tell us the exact ranges of kernel widths. When $\Delta=0$, 
$\gamma > \frac{\lambda^2}{4\mu \pt p}$, meaning a small enough kernel width will always result in oscillatory behavior. When $\Delta \neq 0$, we have $\gamma \in \frac{\lambda^2}{\Delta^2 \pt \mu}(2p-\Delta \pm 2p\sqrt{1-\Delta/p})$, meaning extremely small or extremely large kernel widths will have no oscillations. Lastly for the right plot (\Cref{fig:phase_transition}c), we highlight the range of kernel widths that for a given regularization strength do not affect the convergence rate (Remark \ref{rem:saturation}). For positive $\Delta$ (more target mass than generated), this region intuitively begins where $\rho_a=\rho_b$: as kernel width shrinks further, the magnitude of both $\rho_b$ and $\rho_c$ shrink, leaving $\rho_a$ fixed and dominating. 


In \Cref{fig:linearized_2point}, we observe that our approximation matches true training dynamics very precisely when the learning rate is small. Note that in the small learning rate regime, \cref{cor:etad-constraint} correctly predicts stability for all simulations. In this setting, decreasing the kernel width and increasing regularization can speed up the convergence of the generated point. However, when regularization is small, the effect of kernel width is negligible, as predicted by the large saturation region in \Cref{fig:phase_transition}.

\section{Conclusion}

In this paper we consider a stylized analysis of GAN training using gradient descent ascent. We assumed that the generator was unconstrained and the discriminator was a kernel model (or equivalently a wide neural network in the kernel regime). The analysis uncovers the role of (i) kernel width (or equivalently network depth), (ii) regularization, and (iii) learning rates of generator and discriminator, on the stability and local convergence rate of the dynamics. 

\newpage
\onecolumn
\bibliography{aux/ref}
\bibliographystyle{plain}

\appendix
\section{Linearized System Eigenvalues}

\subsection{Proof of \Cref{thm:discrete}}\label{sec:linearized-rate-proof}
Here we use that $f(x)=f(x;\theta)=\inner{\theta,a(x)}_{\Hilbert'}$ for some feature map $a$ corresponding to the RKHS, which we only need for analytical purposes. Note such a decomposition exists for any RKHS: e.g. we can use a canonical feature map $a(x)=K(x,\cdot)$ corresponding to $\Hilbert'=\Hilbert.$ Similarly, we use $\theta_i:=P_i\theta$, where $P_i$ is the linear map that projects $\theta$ onto the span($\set{a(x)\mid x\in V_i}$). Next, denoting $H(\xt,\thetavec_i)$ as the Hessian of the discriminator
\begin{equation} \label{eq:Hesdef}
    H(\x,\thetavec_i) := 
    \frac{\partial^2 f(\x,\thetavec_i)}{\partial \x^2}\in\Real^{d\times d},
\end{equation}
we first utilize a lemma from \citep{becker2022instability}.
\begin{lemma}[Lemma 4, Becker et al. 2022] \label{lem:evals}
Let  $\z^* = (\thetavec_i^*,\Xt^*_i)$ be an equilibrium
point and let $\Gamma(\z^*)$
be the spectrum of the Jacobian of the update map.
Then $\rho \in \Gamma(\z^*)$ if and only if
$\rho$ is of the form
\begin{equation} \label{eq:rhoetas}
    \rho = 1 + \eta_d s,
\end{equation}
where $s$ is the root of the polynomial:
\begin{align}
    \MoveEqLeft \mathrm{det}(s\I - \A) 
    = (s+\lambda)^{p-Nd}
    \mathrm{det}\left((s+\lambda)(s\I_{Nd} + \Q) 
    + \R\right) 
\end{align}
and where $\Q$ and $\R$ are the block matrices 
with components
\begin{equation} \label{eq:QRdef}
    \Q_{ij} = -\mu \pt_i H(\xt_j^*,\theta^*)\delta_{ij}, 
    \quad
    \R_{ij} = \mu \pt_i \pt_j \frac{\partial^2}{\partial \x \partial \x'} \left. K(\x,\x') \right|_{\x=\xt_i^*,\x'=\xt_j^*}.
\end{equation}
\end{lemma}
When the kernel is an RBF, we can simplify expressions for $\Q$ and $\R$. First, we note that the value of the hessian of the optimal discriminator becomes:
\begin{align}
     H(\x_i, \theta_i^*) &= \left. \frac{\Delta_i}{\lambda}\frac{\partial^2}{\partial\x^2}K(\x,\x_i)\right|_{\x=\x_i}
    = \left. \frac{\Delta_i}{\lambda\sigma^2}\left(\frac{(\x-\x_i)(\x-\x_i)^\intercal}{\sigma^2}-\I_d\right)e^{\norm{\x-\x_i}/2\sigma^2}\right|_{\x=\x_i} =-\frac{\Delta_i}{\lambda\sigma^2}\I_d
\end{align}
Therefore, $\Q$ can now be written as:
$    Q = \frac{\mu\Delta_i}{\lambda\sigma^2}\text{diag}(\pt_1\I_d, \dots, \pt_N\I_d)
$
Next we have:
$
     \left. \frac{\partial^2}{\partial\x\partial\xt}K(\x,\x_i)\right|_{\x=\x'} 
    = \left. \frac{\Delta_i}{\lambda\sigma^2}\left(\frac{(\x-\x_i)(\x-\x_i)^\intercal}{\sigma^2}+\I_d\right)e^{\norm{\x-\x_i}/2\sigma^2}\right|_{\x=\x_i}
    =\frac{1}{\sigma^2}\I_d
$
Therefore the expression for $\R$ becomes:
\begin{equation}
    \R_{jk} = \frac{\mu\pt_j\pt_k}{\sigma^2}\I_d
\end{equation}
\textbf{Single Point Case:} In the simplest case we assume $N=1$. This means the $D(s)$ term can be written as a scalar times a diagonal matrix as follows:
\begin{align}\label{eq:simplest-eigen}
    D(s) &= \left((s+\lambda)(s+\frac{\mu\pt\Delta_i}{\lambda \sigma^2}) + \frac{\mu \pt^2}{\sigma^2}\right)\I_d 
    = \left(s^2 + (\lambda +\frac{\mu\pt\Delta_i}{\lambda \sigma^2})s + \frac{\mu\pt\Delta_i}{\sigma^2} + \frac{\mu \pt^2}{\sigma^2}\right)\I_d 
    = \left(s^2 + (\lambda +\frac{\mu\pt\Delta_i}{\lambda \sigma^2})s + \frac{\mu\pt p_i}{\sigma^2}\right)\I_d
\end{align}
Therefore $\mathrm{det}(D(s))$ becomes
$    \mathrm{det}(D(s)) =\left(s^2 + (\lambda +\frac{\mu\pt\Delta_i}{\lambda \sigma^2})s + \frac{\mu\pt p_i}{\sigma^2}\right)^d 
$
Solving for the roots of this polynomial we have:
\begin{align}
    s = -\frac{1}{2}(\lambda +\frac{\mu\pt\Delta_i}{\lambda \sigma^2}) \pm \frac{1}{2}\sqrt{(\lambda +\frac{\mu\pt\Delta_i}{\lambda \sigma^2})^2-4\frac{\mu\pt p_i}{\sigma^2}}
\end{align}
When $\Delta_i=0$, the eigenvalues simplify to
$
    s = \frac{1}{2}\left(-\lambda\pm \sqrt{\lambda^2-4\frac{\mu \pt^2}{\sigma^2}}\right).
$
Plugging $s$ back into the expression for $\rho$, we see that some of the jacobian eigenvalues must be:
$
    \rho = 1 - \frac{1}{2}\eta_d\lambda \left(1\pm\sqrt{1-4\frac{\mu\pt^2}{\lambda^2\sigma^2}}\right)
$

\textbf{Multi-point Case:} When $N>1$, we can write $D(s)$ as the following block matrix:
\begin{equation}
    D(s)_{jk} = \begin{cases} \left[(s+\lambda)(s+\frac{\mu \pt_j\Delta}{\lambda \sigma^2})+\frac{\mu \pt_j^2}{\sigma^2}\right] \I_d & j=k \\
    \left[\frac{\mu \pt_j\pt_k}{\sigma^2} \right]\I_d & j\neq k
    \end{cases}
\end{equation}
Note that because each block is diagonal and of the same dimension, their matrix products are commutative. Therefore, we can make use of the following theorem in \citep{silvester2000determinants}:
\begin{theorem}[Theorem 1, Silvester 2000]\label{thm:det-det}
Let R be a commutative subring of ${}^nF^n$, where F is a field (or a commutative ring), and let$M\in{}^mR^m$. Then
\begin{equation}
    \mathrm{det}_F(M) = \mathrm{det}_F(\mathrm{det}_R(M))
\end{equation}
\end{theorem}
Let's assume that $\pt_j=\pt_k \: \forall j,k\in V_0$. This means that our block matrix is of the form:
$D(s) = \begin{bmatrix}
        D_1 & D_2 & \cdots & D_2 \\
        D_2 & D_1 & \cdots & D_2 \\
        D_2 & D_2 & \cdots & D_1 
        \end{bmatrix} = (D_1-D_2)(\I+\U\V^\intercal)$
Where $U^\intercal=(D_2,D_2 \dots)$, and $V^\intercal=((D_1-D_2)^{-1},(D_1-D_2)^{-1} \dots)$. Therefore by \cref{thm:det-det} and the matrix determinant lemma, we can write the determinant as:
\begin{align}
    \mathrm{det}(D(s)) &= \mathrm{det}\left((D_1-D_2)^N(\I_d + ND_1(D_2-D_1)^{-1}) \right) = \mathrm{det}\left(\left[(s+\lambda)(s+\frac{\mu\pt\Delta}{\lambda\sigma^2})\right]^N\left[1+N\frac{\mu\pt^2/\sigma^2}{(s+\lambda)(s+\frac{\mu\pt\Delta}{\lambda\sigma^2})}\right]\I_d\right) \nonumber \\
    &=\left[(s+\lambda)(s+\frac{\mu\pt\Delta}{\lambda\sigma^2})\right]^{Nd}\left[1+N\frac{\mu\pt^2/\sigma^2}{(s+\lambda)(s+\frac{\mu\pt\Delta}{\lambda\sigma^2})}\right]^d= \left[(s+\lambda)(s+\frac{\mu\pt\Delta}{\lambda\sigma^2})\right]^{(N-1)d}\left[(s+\lambda)(s+\frac{\mu\pt\Delta}{\lambda\sigma^2})+N\frac{\mu\pt^2}{\sigma^2}\right]^d\nonumber \\
    &= \left[(s+\lambda)(s+\frac{\mu\pt\Delta}{\lambda\sigma^2})\right]^{(N-1)d}\left[s^2 + (\lambda +\frac{\mu\pt\Delta_i}{\lambda \sigma^2})s + \frac{\mu\pt p_i}{\sigma^2}\right]^d = \left[(s+\lambda)(s+b)\right]^{(N-1)d}\left[s^2 + (\lambda +b)s + c\right]^d,
\end{align}
where, in the last step, we used the expressions for $b$
and $c$ in \eqref{eq:abcdef}.
The roots of the third polynomial are
$s = -\frac{\lambda + b}{2} \pm \sqrt{ \left(\frac{\lambda + b}{2}
    \right)^2 - c  }.$
Note that when $N=1$ this polynomial reduces to \cref{eq:simplest-eigen}. When $N>1$, the main difference is that we introduce a new root at $s=-\frac{\mu\pt\Delta}{\lambda\sigma^2}$. The full characteristic polynomial for matrix A can now be written as:
\begin{equation}
    \mathrm{det}(s\I-\A) = (s+\lambda)^{p-d}(s+\frac{\mu\pt\Delta}{\lambda\sigma^2})^{(N-1)d}\left[s^2 + (\lambda +\frac{\mu\pt\Delta_i}{\lambda \sigma^2})s + \frac{\mu\pt p_i}{\sigma^2}\right]^d
\end{equation}
The polynomial above has roots $a,b,$ or $c$, which proves the claim of Theorem \cref{thm:discrete}.
\subsection{Proof of Corollary \ref{cor:etad-constraint} and Proposition \ref{prop:linear-sufficient-condition}}
From \Cref{thm:discrete}, we know that the eigenvalues of the linearized RBF Kernel GAN around the true point can be written as:
$\rho = \left\{1-\eta_d a,\: 1-\eta_d b,\: \left(1 -\eta_d\frac{(a+b)}{2}\right) \pm \eta_d\sqrt{\left[\frac{(a+b)}{2}\right]^2-c}\right\}
$.
Given that $|\rho|<1$ is a necessary condition for stability, we can derive some constraints on our hyperparameters. We assume here that $\lambda$, $\sigma$, and $\eta_d, \eta_g$ are all positive values.\\
\textbf{Stability of $\rho_a$ and $\rho_b$:} From the first two eigenvalues we have the condition:
$a \in (0,2/\eta_d), b \in (0,2/\eta_d)$
It is easy to see that we arrive at the conditions $\lambda < 2/\eta_d$ and $\Delta >0$. From the second bound we can substitute $\pt = (\Delta-\p_0)/N$ and solve a quadratic equation for bounds on $\Delta$:
\begin{align}
    \frac{\eta_g\pt\Delta}{\lambda \sigma^2} < 2 &\implies \eta_g(\frac{\Delta-\p_0}{N})\Delta <2\lambda\sigma^2 \implies \frac{\eta_g}{N} \Delta^2 + \frac{\eta_g\p_0}{N}\Delta-2\lambda\sigma^2 <0  
\end{align}
The roots of this quadratic are:
\begin{align}
    \Delta &= \frac{N}{2\eta_g}\left(-\frac{\eta_g\p_0}{N}\pm \sqrt{\frac{\eta_g^2\p_0^2}{N^2}+\frac{8\eta_g\lambda\sigma^2}{N}}\right) = -\frac{1}{2}\left(\p_0 \pm \p_0 \sqrt{1+\frac{8N\lambda\sigma^2}{\eta_g\p_0^2}}\right) =-\frac{\p_0}{2}\left(1 \pm \sqrt{1+\frac{8N\lambda\sigma^2}{\eta_g\p_0^2}}\right)
\end{align}
Therefore we have the additional bound on $\Delta$:
$\Delta \in \left[0, \frac{\p_0}{2}\left(-1 + \sqrt{1+\frac{8N\lambda\sigma^2}{\eta_g\p_0^2}}\right)\right]
$
The interval becomes $\Delta \in [0, \p_0]$ only if 
\begin{equation}
    \sqrt{1+\frac{8N\lambda\sigma^2}{\eta_g\p_0^2}} > 3 \\
    \implies \frac{N\lambda \sigma^2}{\eta_g} > \p_0^2
\end{equation}
It is sufficient to have $\lambda>\frac{\eta_g}{N\sigma^2}$ for this condition to hold. \\
\textbf{Stability of $\rho_c$:} Let us denote the quantity $\alpha:= \eta_d(a+b)/2=  (\eta_d\lambda +\frac{\eta_g\pt\Delta_i}{\lambda \sigma^2})/2$, and $\beta:=\eta_d^2c = \frac{\eta_d\eta_g\pt p_i}{\sigma^2}$. We know that when the previous conditions are satisfied $\alpha \in [0,2]$. We can write the third eigenvalue pair as:
\begin{equation}
    \rho = (1-\alpha)\pm\sqrt{\alpha^2-\beta}
\end{equation}
We know that $\beta > 0$ already because each component is positive. The last way these eigenvalues could have magnitude larger than 1 is if they have large imaginary components. The following constraint ensures these imaginary components are not too large:
\begin{align}
    (1-\alpha)^2 - \alpha^2 +\beta < 1 &\implies 1-2\alpha + \beta < 1 \implies \beta < 2\alpha \implies c < (a+b)/\eta_d
\end{align}
Using the inequalities $\pt p_0 <1$, $\Delta>0$, we get the last sufficient condition for our proposition:
\begin{equation}
    \lambda > \frac{\eta_g}{\sigma^2} \implies \frac{\eta_g\pt\p_0}{\sigma^2} < \lambda +  \frac{\mu\pt\Delta}{\lambda^2} \Leftrightarrow c < (a+b)/\eta_d
\end{equation}

\begin{figure*}
\centering\includegraphics[width=0.9\linewidth]{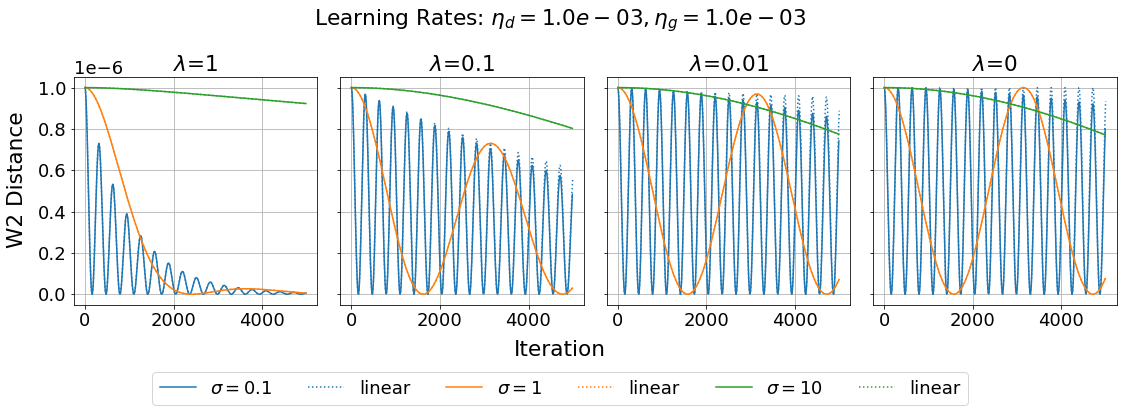}
    \caption{GAN training behavior with RBF kernel discriminator. One generated point is initialized close to the target, in order to compare predicted local convergence rates (dotted lines) to empirical convergence (solid lines).}
    \label{fig:linearized_2point}
\end{figure*}
\end{document}